\documentclass[letterpaper,10pt,conference]{ieeeconf}

\IEEEoverridecommandlockouts
\usepackage{graphicx}        %
\usepackage{amsmath,amssymb} %
\usepackage{booktabs}        %
\usepackage{multirow}        %
\usepackage{tabularx}        %
\usepackage{adjustbox}       %
\usepackage{xcolor}          %
\usepackage{xspace}          %
\usepackage{cite}            %
\usepackage{url}             %
\usepackage{microtype}
\usepackage[english]{babel}
\usepackage{needspace}

\renewcommand{\thetable}{\arabic{table}}
\makeatletter
\renewcommand{\fnum@table}{Table~\thetable}
\let\dkwbc@originalmakecaption\@makecaption
\long\def\@makecaption#1#2{%
    \ifx\@captype\@IEEEtablestring
        \@IEEEtablecaptionsepspace
        {\normalsize\normalfont\noindent\textbf{#1.}\enspace #2\par}%
    \else
        \begingroup
        \let\footnotesize\normalsize
        \dkwbc@originalmakecaption{#1}{#2}%
        \endgroup
    \fi
}
\makeatother

\newcommand{\method}{\textsc{DKWBC}\xspace}

\newcommand{\reffig}[1]{Fig.~\ref{#1}}
\newcommand{\reftab}[1]{Table~\ref{#1}}

\usepackage[hidelinks]{hyperref}
\usepackage[capitalize]{cleveref}
\hypersetup{
    pdftitle={DKWBC: Decoupling Motion Tracking and Balancing for Humanoid Whole-Body Control},
    pdfauthor={Dingkun Wei, Zehong Shen, Huaijin Pi, Xueyang Wang, Xing Zhu, Hujun Bao, Yujun Shen, Xiaowei Zhou},
    pdfsubject={Humanoid whole-body control},
    pdfkeywords={humanoid robots, whole-body control, inverse dynamics, residual learning}
}

\title{\LARGE \bf
\method: Decoupling Motion Tracking and Balancing for Humanoid Whole-Body Control
}

\author{Dingkun Wei$^{1,2}$, Zehong Shen$^{2,*}$, Huaijin Pi$^{2,3}$, Xueyang Wang$^{1,2}$,\\
Xing Zhu$^{2}$, Hujun Bao$^{1}$, Yujun Shen$^{2,\dagger}$, Xiaowei Zhou$^{1,\dagger}$\\[0.25em]
{\small $^{*}$Project Lead \quad $^{\dagger}$Corresponding Authors}\\[0.25em]
{\small $^{1}$Zhejiang University \quad $^{2}$Robbyant \quad $^{3}$The University of Hong Kong}%
\thanks{Dingkun Wei, Huaijin Pi, and Xueyang Wang conducted this work during their internships at Ant Group.
This work was supported by Ant Group Research Intern Program.}}
\date{}

\begin{document}

\maketitle
\suppressfloats[t]
\thispagestyle{plain}
\pagestyle{plain}
\raggedbottom

\begin{abstract}
Humanoid whole-body control requires a robot to track diverse reference motions while maintaining balance under gravity, torque limits, contact switching, and external disturbances.
Existing reinforcement-learning-based trackers typically learn control outputs that jointly address motion tracking and balance.
However, similar reference poses can correspond to substantially different dynamic demands, requiring a unified policy to learn motion-dependent control requirements alongside balance corrections.
This can complicate learning and lead to high-frequency fluctuations in control outputs.
We propose \method, a dynamics-informed whole-body control framework that combines explicit model-based tracking with a learned balancing residual.
Inspired by the model-based and environmental-wrench contributions to joint torques in floating-base dynamics, \method uses inverse-dynamics feedforward and joint PD feedback to supply tracking torques, while a residual policy learns additional joint-torque corrections for balanced execution.
By explicitly accounting for motion-dependent torque demands, this structure simplifies the policy's learning task.
Experiments in simulation and on real hardware demonstrate that a single learned balancing policy generalizes across motion datasets while maintaining accurate motion tracking and balance across ground support, attached payloads, and physical interaction.
Project page: \url{https://robbyant-research.github.io/DKWBC/}.
\end{abstract}

\section{Introduction}
\label{sec:intro}

 \begin{figure}[t]
      \centering
      \includegraphics[width=\columnwidth]{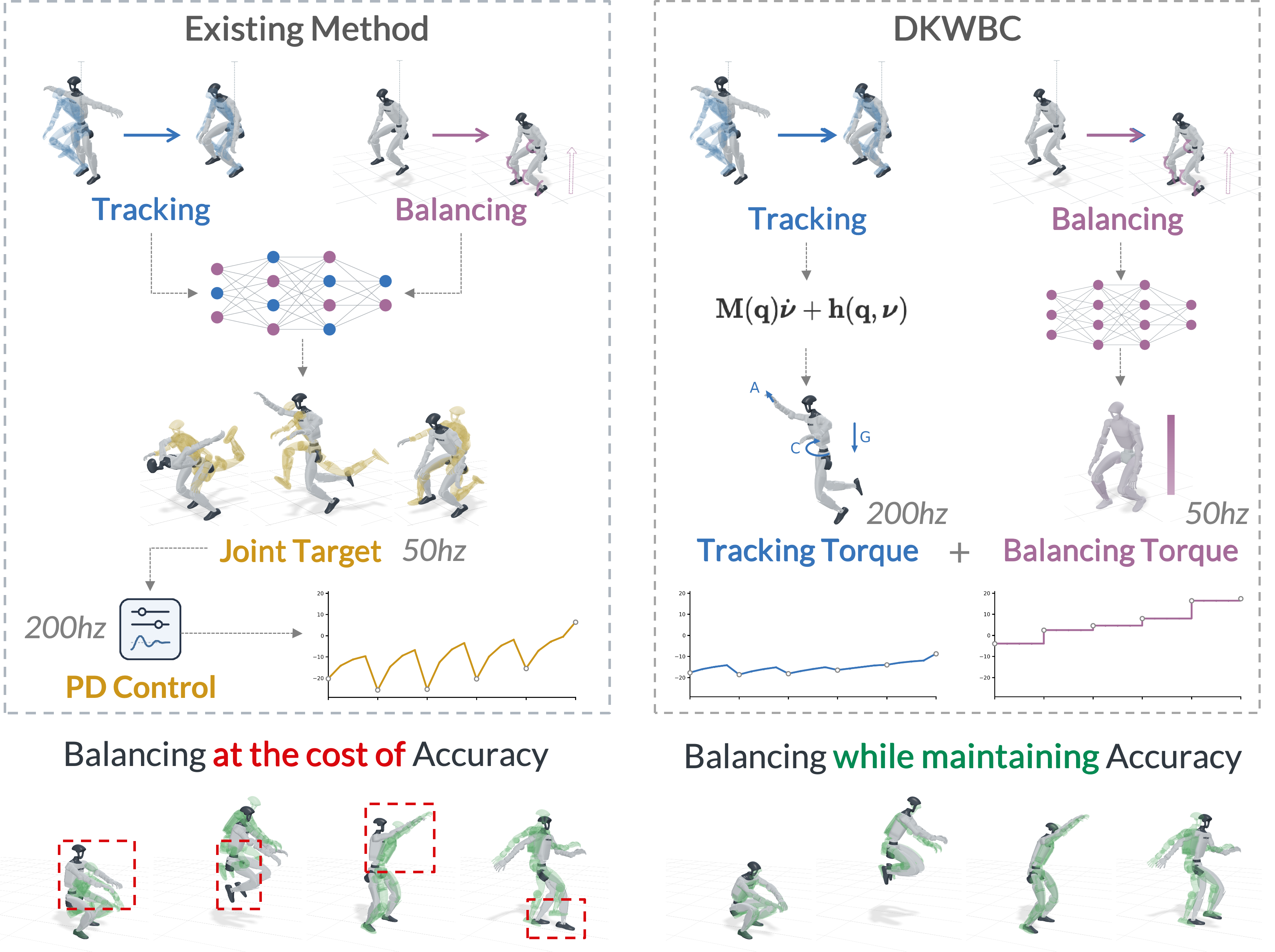}
      \caption{Decoupling motion tracking and balancing.
      Left: A conventional policy jointly learns tracking and balancing, producing joint position targets that a proportional-derivative (PD) controller converts into torques.
      Right: \method combines nominal tracking torques computed by inverse dynamics with a learned balancing residual for balance.
      Bottom: On the unseen motions shown, \method tracks the reference poses more closely while maintaining balance.}
      \label{fig:intro_single}
  \end{figure}
Humanoid robots offer a promising platform for general-purpose embodied intelligence, with the potential to move, manipulate, and interact flexibly with the physical world~\cite{sentis2006}.
Realizing this potential requires a single whole-body control policy that enables accurate and stable execution of diverse motions.
Recent studies have advanced humanoid whole-body control by scaling human motion datasets~\cite{sonic}, developing more expressive policy architectures~\cite{gmt,humanoidgpt}, and learning richer representations of motion and dynamics~\cite{unitracker,bfmzero,any2track}.

Despite these recent advances, many approaches rely on a policy to predict joint position targets that a proportional-derivative (PD) controller converts into motor torques, jointly addressing motion tracking and balance~\cite{beyondmimic,sonic}.
Diverse motions, however, impose different dynamic demands: even similar reference poses can be associated with substantially different velocities, accelerations, and required torques.
A unified policy must learn these motion-dependent control requirements while maintaining balance under changing contacts and external loads.
These competing demands can complicate learning and may encourage conservative outputs, reducing tracking accuracy across diverse motions.

We propose \method, a whole-body control framework that combines model-based motion tracking with learned residual control.
Its core idea is inspired by the floating-base equations of motion \cite{herzog2016,gams2018}, where joint torques can be expressed in terms of a model-based contribution determined by the robot state and acceleration and a contribution from environmental wrenches.
As shown in \cref{fig:intro_single}, \method consists of a model-based tracking module that supplies nominal tracking torques and a residual policy that provides additional torque corrections for balanced execution.
Specifically, the tracking module computes inverse-dynamics feedforward torques from the current state and reference acceleration, supplemented by joint PD feedback, while the residual policy learns additional joint torques through interaction with the environment.
By explicitly modeling motion-dependent torque demands, this separation simplifies the policy's learning task: learning residual corrections for balanced execution under the supplied tracking torques.

This shift in the learning task enables \method to outperform SONIC~\cite{sonic}, a strong whole-body motion tracker, in scaling experiments under comparable training settings.
On both BONES-SEED test sets, \method achieves higher tracking accuracy and success rates than retrained SONIC across all evaluated training-data scales.
Furthermore, by focusing learning on balanced execution, \method maintains accurate motion tracking and balance under attached payloads and external disturbances, as demonstrated in simulation and real-robot tests.
Our contributions are threefold:
\begin{enumerate}
    \item We propose \method, a whole-body control framework that decouples motion tracking from balancing control by combining model-based nominal tracking torques with a learned balancing residual to maintain balance while accurately tracking reference motions.
    \item We design the two modules around their distinct objectives: inverse dynamics and joint feedback provide accurate motion tracking, while feedforward-torque observations and direct residual torque actions encourage the policy to learn balancing capabilities shared across motions.
    \item We demonstrate that a single balancing policy generalizes across motion datasets while maintaining accurate motion tracking and balance across ground support, attached payloads, and external interaction, with validation in simulation and on real hardware.
\end{enumerate}

\section{Related Work}
\label{sec:related}

\textbf{Humanoid whole-body motion tracking.}
Given a reference robot motion, a motion-tracking policy computes actuator commands to reproduce the reference motion.
DeepMimic~\cite{deepmimic} established motion imitation for learning simulated skills, while BeyondMimic~\cite{beyondmimic} combines motion tracking on real humanoids with guided diffusion for skill composition.
SONIC~\cite{sonic} scales motion data and policy capacity, while GMT~\cite{gmt} and Humanoid-GPT~\cite{humanoidgpt} use mixture-of-experts and causal Transformer policies, respectively.
Representation learning further supports general control: UniTracker~\cite{unitracker} encodes motion with a conditional variational autoencoder, BFM-Zero~\cite{bfmzero} represents motions, goals, and rewards in a shared latent space, and Any2Track~\cite{any2track} learns dynamics embeddings through world-model prediction for online adaptation.
Our work complements these advances by computing motion-tracking torques with inverse dynamics and learning a support residual, changing the physical role of the policy output.

\textbf{Model-based and hybrid whole-body control.}
Classical whole-body controllers use robot dynamics to compute joint torques that track desired motions while maintaining balance and satisfying contact constraints, such as keeping supporting feet from slipping~\cite{sentis2006,herzog2016}.
Learning has long been used to improve these controllers: Gams et al. learn corrections to the desired center-of-mass acceleration within inverse-dynamics control~\cite{gams2018}, 
SEEC combines model-based end-effector tracking with model-guided residual learning for arm stabilization, including transfer to unseen locomotion controllers~\cite{seec}.
\method instead decomposes the torques at all actuated joints: inverse dynamics computes the motion-tracking torque at execution time, while a learned support residual handles support and external loading without an online contact-force optimization.
The residual can also compensate for model errors; the decomposition alone does not identify physical contact forces.

\textbf{Residual tracking and policy adaptation.}
Residual methods differ in both their baseline and the quantity being corrected.
HiLo~\cite{hilo} and RobotDancing~\cite{robotdancing} learn joint-target corrections around reference motions.
ASAP learns a delta-action model from real rollouts and inserts it into the simulator for policy fine-tuning~\cite{asap}.
OmniXtreme uses residual post-training with actuator-aware modeling for highly dynamic skills~\cite{omnixtreme}.
FADA separates a learned planner from a neural inverse-dynamics module and adapts the latter with target-domain rollouts~\cite{fada}.
We train the residual alongside an analytic tracking component that computes the motion-tracking torque by inverse dynamics at execution time.
The distinction is therefore the torque-level allocation of motion tracking and support control, rather than the use of residual learning alone.

\textbf{Force adaptation and physical interaction.}
FALCON~\cite{falcon} and HAFO~\cite{hafo} coordinate upper- and lower-body policies for force-adaptive loco-manipulation; FALCON uses inverse dynamics to construct its torque-aware training curriculum.
ResMimic refines a frozen general motion tracker with task-specific residuals for object interaction~\cite{resmimic}, whereas SceneBot conditions a general tracker on per-link contact labels to support terrain and object interactions~\cite{scenebot}.
CHIP enables controllable end-effector compliance through hindsight perturbation~\cite{chip}, while ADAPT supplies a locomotion policy with analytical disturbance-observer estimates~\cite{adaptobserver}.
The YAHMP study further examines how hand-force randomization and action parameterization affect tracking and physical interaction~\cite{yahmp}.
We evaluate full-body motion tracking and balance across ground support, attached payloads, and interaction forces using one support residual and run-time inverse-dynamics feedforward.
This evaluation does not establish explicit force regulation or manipulation-task success.

\section{Method}
\label{sec:method}

\begin{figure*}[t]
    \centering
    \includegraphics[width=0.87\textwidth]{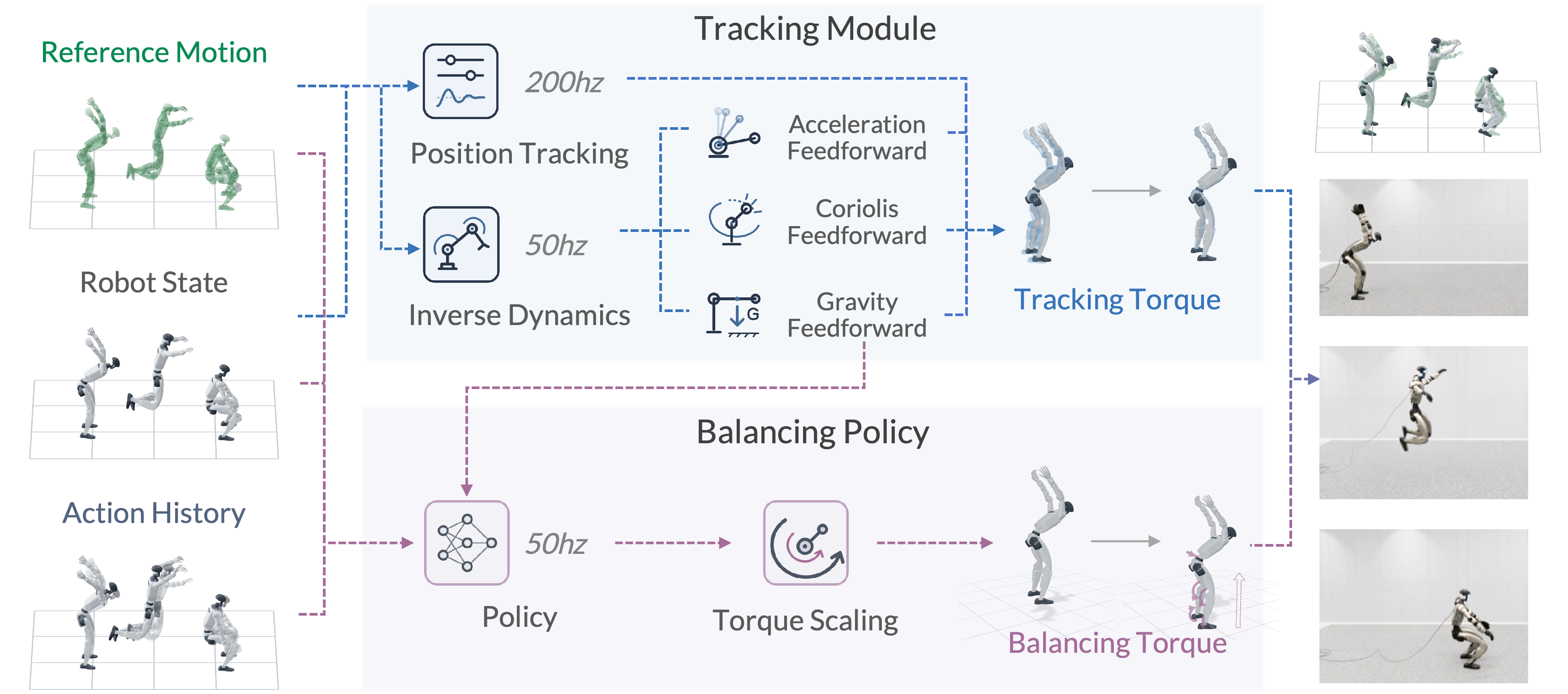}
    \caption{Overview of \method.
    The model-based tracking module combines inverse-dynamics feedforward with joint feedback.
    The residual policy receives robot and reference information together with the feedforward components, and predicts joint-torque corrections for balanced motion execution.
    The tracking and residual torques are summed to form the final joint-torque command, enabling the real robot to perform challenging motions such as standing long jumps.}
    \label{fig:method-overview}
\end{figure*}

Our goal is to accurately track diverse reference motions while maintaining balance.
To this end, we present \method, which combines model-based tracking with learned residual control.
As shown in \cref{fig:method-overview}, \method consists of an inverse-dynamic based tracking module and a balancing policy.
Section~\ref{sec:method-dynamics} explains the dynamics underlying our control decomposition, Section~\ref{sec:method-tracking} describes the tracking module, and Section~\ref{sec:method-balance} defines the balancing policy and its learning objective.

\subsection{Dynamics-Informed Control Decomposition}
\label{sec:method-dynamics}

Given a floating-base humanoid with $n$ actuated joints,
we describe its configuration by $\mathbf q=(\mathbf p,\mathbf R,\boldsymbol\theta)\in\mathbb R^3\times\mathrm{SO}(3)\times\mathbb R^n$, and its velocity by $\boldsymbol\nu=(\mathbf v,\boldsymbol\omega,\dot{\boldsymbol\theta})\in\mathbb R^{6+n}$.
Here, $\mathbf p$ and $\mathbf R$ describe the base position and orientation, $\mathbf v$ and $\boldsymbol\omega$ are the world-frame base linear and angular velocities, and $\boldsymbol\theta$ contains the joint positions.
The equations of motion take the standard form~\cite{herzog2016,gams2018}:

\begin{equation}
\label{eq:method-dynamics}
\mathbf M(\mathbf q)\dot{\boldsymbol\nu}
+\mathbf h(\mathbf q,\boldsymbol\nu)
=\mathbf S^\top\boldsymbol\tau+\sum_k\mathbf J_k^\top\boldsymbol\lambda_k.
\end{equation}

Here, $\mathbf M$ is the inertia matrix, including reflected actuator inertia, and $\mathbf h$ contains Coriolis, centrifugal, and gravity terms.
The vector $\boldsymbol\tau\in\mathbb R^n$ contains the actuated joint torques, and $\mathbf S=[\mathbf 0_{n\times6}\;\mathbf I_n]$ selects the joint coordinates.
Each external wrench $\boldsymbol\lambda_k$ acts on the robot through the corresponding Jacobian $\mathbf J_k$.
Multiplying Eq.~\eqref{eq:method-dynamics} by $\mathbf S$ and rearranging yields:

\begin{equation}
\label{eq:method-actuated-dynamics}
\boldsymbol\tau
=\mathbf S\!\left[\mathbf M(\mathbf q)\dot{\boldsymbol\nu}+\mathbf h(\mathbf q,\boldsymbol\nu)\right]
-\mathbf S\sum_k\mathbf J_k^\top\boldsymbol\lambda_k.
\end{equation}

Equation~\eqref{eq:method-actuated-dynamics} reveals two contributions to the joint torque: a model-based term that can be computed given the robot state and acceleration, and an environmental wrench term involving contact and external forces that are difficult to observe directly.
This motivates a control structure with two modules: a model-based tracking module that explicitly computes nominal tracking torques, and a residual policy that learns additional torque corrections for balanced execution.

The two modules have distinct objectives and challenges.
The tracking module aims to follow the reference motion as accurately as possible.
A key challenge is that reference motions retargeted from humans are kinematically feasible but not necessarily dynamically feasible for the robot.
The balancing module uses reinforcement learning to maintain balance while the robot executes the motion driven by the supplied tracking torques.
Within the motion-imitation RL framework, its design should discourage the policy from relying on motion-specific tracking compensation, which can limit generalization across motions.
Sections~\ref{sec:method-tracking} and~\ref{sec:method-balance} describe how we design the two modules around these objectives and address their respective challenges.

\subsection{Model-Based Tracking Torque}
\label{sec:method-tracking}

The tracking module evaluates the model-based term using the current robot state and the reference acceleration:
\begin{equation}
\label{eq:method-nominal-feedforward}
\boldsymbol\tau_{\mathrm{id}}
=\mathbf S\!\left[
\widehat{\mathbf M}(\mathbf q)\dot{\boldsymbol\nu}^{\mathrm{ref}}
+\widehat{\mathbf h}(\mathbf q,\boldsymbol\nu)
\right],
\end{equation}
where $\widehat{\mathbf M}$ and $\widehat{\mathbf h}$ are the nominal model quantities, and $\dot{\boldsymbol\nu}^{\mathrm{ref}}$ is the reference acceleration.

\textbf{Reference derivatives.}
The tracking module takes the current state $(\mathbf q,\boldsymbol\nu)$ and the reference motion $\mathbf q^{\mathrm{ref}}(t)$ as inputs.
Reference joint velocities and base linear velocities are obtained by finite differences, and reference base angular velocities from relative rotations on $\mathrm{SO}(3)$.
We smooth the reference velocities with a three-frame Gaussian filter ($\sigma=0.75$ frames) and differentiate them to obtain $\dot{\boldsymbol\nu}^{\mathrm{ref}}$.

\textbf{Inverse-dynamics feedforward.}
We evaluate Eq.~\eqref{eq:method-nominal-feedforward} with the recursive Newton--Euler algorithm~\cite{rnea} on the floating-base model, where gravity enters through the base acceleration and reflected actuator inertia is included in $\widehat{\mathbf M}$.
Writing $\widehat{\mathbf h}(\mathbf q,\boldsymbol\nu)=\widehat{\mathbf c}(\mathbf q,\boldsymbol\nu)+\widehat{\mathbf g}(\mathbf q)$ splits the feedforward into three components, which also serve as policy observations in Section~\ref{sec:method-balance}:

\begin{equation}
\label{eq:method-feedforward-components}
\begin{aligned}
\boldsymbol\tau_{\mathrm{acc}}
&=\mathbf S\widehat{\mathbf M}(\mathbf q)\dot{\boldsymbol\nu}^{\mathrm{ref}},\\
\boldsymbol\tau_{\mathrm{cor}}
&=\mathbf S\widehat{\mathbf c}(\mathbf q,\boldsymbol\nu),\\
\boldsymbol\tau_{\mathrm{grav}}
&=\mathbf S\widehat{\mathbf g}(\mathbf q),\\
\boldsymbol\tau_{\mathrm{id}}
&=\boldsymbol\tau_{\mathrm{acc}}+\boldsymbol\tau_{\mathrm{cor}}+\boldsymbol\tau_{\mathrm{grav}}.
\end{aligned}
\end{equation}

Evaluating these terms at the current state captures configuration- and velocity-dependent effects, while $\dot{\boldsymbol\nu}^{\mathrm{ref}}$ specifies the nominal motion demand.

\textbf{Joint feedback.}
With an exact model, inverse dynamics provides the nominal torque for a dynamically feasible reference motion.
As noted in Section~\ref{sec:method-dynamics}, feedforward alone may leave tracking errors for dynamically infeasible references.
We therefore add joint-space PD feedback to correct joint position and velocity errors, using the fixed diagonal gain matrices $\mathbf K_p^{\mathrm{trk}}$ and $\mathbf K_d^{\mathrm{trk}}$ of BeyondMimic~\cite{beyondmimic}.
Although this feedback contributes less than 10\% of the complete tracking torque, it is essential for accurate execution.
The feedback torque is

\begin{equation}
\label{eq:method-feedback}
\begin{aligned}
\boldsymbol\tau_{\mathrm{pd}}
&=\mathbf K_p^{\mathrm{trk}}(\boldsymbol\theta^{\mathrm{ref}}-\boldsymbol\theta)\\
&\quad+\mathbf K_d^{\mathrm{trk}}(\dot{\boldsymbol\theta}^{\mathrm{ref}}-\dot{\boldsymbol\theta}).
\end{aligned}
\end{equation}

Together, $\boldsymbol\tau_{\mathrm{id}}$ and $\boldsymbol\tau_{\mathrm{pd}}$ form the complete tracking torque, which is added to the learned residual:
\begin{equation}
\label{eq:method-command}
\begin{aligned}
\boldsymbol\tau_{\mathrm{trk}}
&=\boldsymbol\tau_{\mathrm{id}}+\boldsymbol\tau_{\mathrm{pd}},\\
\boldsymbol\tau_{\mathrm{cmd}}
&=\boldsymbol\tau_{\mathrm{trk}}+\boldsymbol\tau_{\mathrm{bal}}.
\end{aligned}
\end{equation}

\subsection{Learning Residual Corrections for Balanced Execution}
\label{sec:method-balance}

Reinforcement learning for motion imitation optimizes a control policy through closed-loop interaction between the robot and a simulated environment, using rewards based on reference-motion tracking errors and measures of stability to enable accurate and stable motion execution~\cite{beyondmimic}.
We retain this framework for residual policy learning.
To encourage the policy to learn a generalizable ability to maintain balance under the supplied tracking torques and discourage tracking compensation, we introduce two modifications to its observations and actions.

\textbf{Tracking-torque observations.}
We explicitly expose the nominal tracking demand by adding the three $n$-dimensional feedforward components $\boldsymbol\tau_{\mathrm{acc}}$, $\boldsymbol\tau_{\mathrm{cor}}$, and $\boldsymbol\tau_{\mathrm{grav}}$ from Eq.~\eqref{eq:method-feedforward-components} to the observation.
This lets the policy condition its corrections on the torque already supplied by the tracking module.
The actor otherwise follows BeyondMimic, using a single observation frame that contains the reference-motion command and proprioceptive information, excluding quantities unavailable at deployment.
The last-action feature records the previous combined policy and feedforward torque, $\boldsymbol\tau_{\mathrm{bal},t-1}+\boldsymbol\tau_{\mathrm{id},t-1}$.

\textbf{Direct torque actions.}
A common design in RL-based motion tracking is to predict joint-position targets at a lower frequency (e.g., $50\,\mathrm{Hz}$) and convert them into torques through a higher-frequency PD controller (e.g., $200\,\mathrm{Hz}$).
The policy output thus passes through a position servo: even when the output is held fixed, the torque changes at each PD update with the current joint position and velocity.
\method instead predicts residual joint torques directly at $50\,\mathrm{Hz}$ and holds them constant between policy updates, while the tracking module provides joint feedback around the reference motion.

This design is motivated by two considerations.
First, the policy supplies additional torque directly, rather than indirectly generating torque by adjusting joint-position targets through a position servo.
This removes the pathway through which learned position targets participate in fast joint feedback, making it harder for the policy to use that pathway for motion-specific tracking compensation.
Second, the tracking module already provides nominal tracking torques and joint-error feedback.
We therefore aim for the balancing policy to supply the additional torques required for balanced execution and to learn balancing capabilities shared across motions, rather than duplicating the joint-error corrections already provided by the tracking module.

For stable training, each action is scaled by a fixed fraction of the corresponding torque limit $\tau_j^{\max}$:

\begin{equation}
\boldsymbol\tau_{\mathrm{bal},t}=\mathbf s_\tau\odot\mathbf a_t,
\qquad
s_{\tau,j}=0.25\,\tau_j^{\max},
\label{eq:method-balance-action}
\end{equation}
where $\odot$ denotes elementwise multiplication, so a unit policy output corresponds to 25\% of the joint torque limit.
The residual is added to the tracking torque as in Eq.~\eqref{eq:method-command}.

\section{Experiments}

\label{sec:experiments}

Our evaluation focuses on three aspects:

\begin{enumerate}
    \item Whether the \method{} policy learns more transferable balancing capabilities as the amount of training motion data increases.
    \item Whether \method can adapt to different external forces while maintaining tracking and balance.
    \item Ablations of \method to examine explicit decoupling and the effectiveness of its Tracking Module and Balancing Execution.
\end{enumerate}

\subsection{Training-Data Scaling}

\label{sec:exp_scaling}

\begin{figure*}[tp]
    \centering
    \includegraphics[width=0.76 \textwidth]{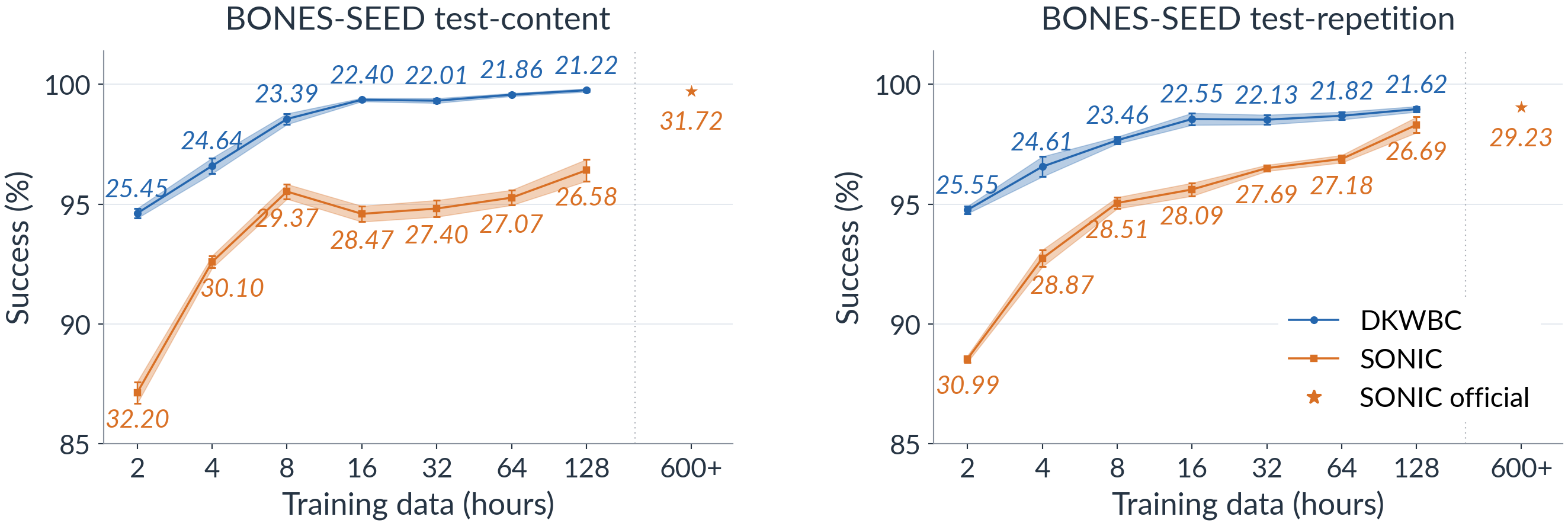}%
    \caption{Motion-tracking performance of \method compared with SONIC across training-data scales.
The left and right panels show results on BONES-SEED test-content and test-repetition~\cite{sonic}, respectively.
Curves and shaded bands show the mean success rate and $\pm1$ sample standard deviation across four evaluation checkpoints per configuration. Labels indicate MPJPE (mm), averaged over successfully tracked motions within each checkpoint and then across checkpoints. Stars mark results from the official SONIC checkpoint.
}
    \label{fig:scaling-success-mpjpe}
\end{figure*}

\begin{figure*}[t]
    \centering
    \includegraphics[width=0.85\textwidth]{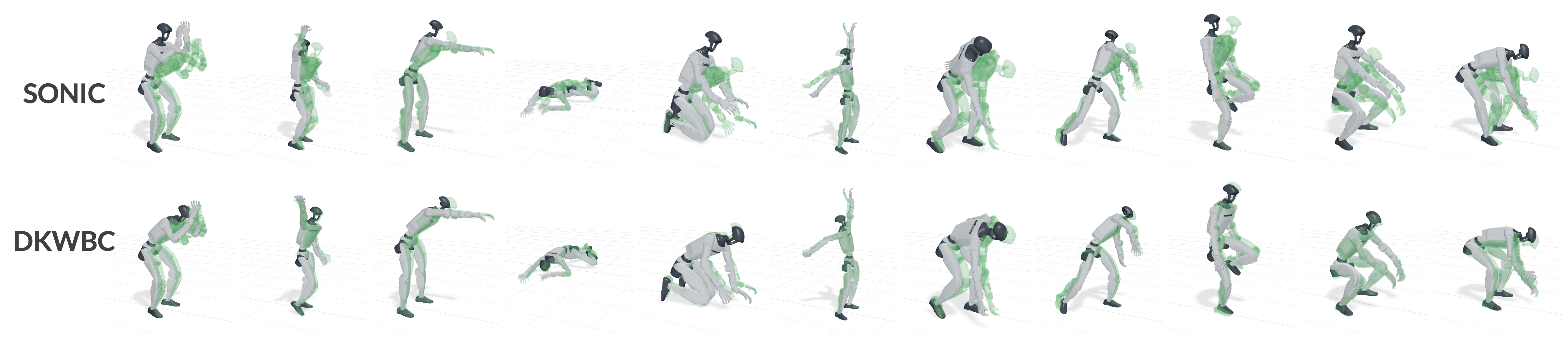}
    \caption{Qualitative motion-tracking comparison between SONIC (top) and \method (bottom).
Solid robots show the executed poses, and translucent green robots show the reference poses.}
    \label{fig:scaling-qualitative}
\end{figure*}

\paragraph{Baseline}
We compare \method with SONIC~\cite{sonic}, a whole-body controller trained on large motion datasets.
We adapt its official training pipeline for G1 motion tracking, removing VR inputs and SMPL parameters from motion-encoder training.

\paragraph{Training Data}
We use the BONES-SEED dataset~\cite{sonic} for training.
We manually remove motions that are infeasible in our setting, such as leaning on external objects or penetrating the ground, and discard mirrored motions.
The resulting dataset contains \textbf{128.53 hours} of motion.
To study the effect of training-data scale, we uniformly sample subsets of 2, 4, 8, 16, 32, 64, and 128 hours from this dataset for training.

\paragraph{Evaluation Protocol}
We evaluate success rate and MPJPE on BONES-SEED test-content and test-repetition~\cite{sonic}.
Following the thresholds in the official SONIC repository, a motion is considered successful only if every frame satisfies all three conditions: pelvis height error does not exceed 0.25 m, pelvis orientation error does not exceed 1 rad, and the height error of each wrist and ankle does not exceed 0.25 m.
For each checkpoint, success rate is the proportion of successfully tracked motions in the full test set, and MPJPE is averaged over those successful motions with equal weight per motion.
We report the mean success rate and MPJPE across four checkpoints from the same training run for each configuration, with shaded bands indicating $\pm1$ sample standard deviation of the success rate.
Test-content contains motions from categories unseen during training, while test-repetition contains unseen motions from categories seen during training.
We manually exclude infeasible motions, such as floating and sitting on a bench, and then remove mirrored motions.
BONES-SEED test-content contains \textbf{3,145 motions totaling 6.031 hours}, and BONES-SEED test-repetition contains \textbf{3,019 motions totaling 6.002 hours}.
For a fair comparison, \method{} uses the same training-related settings as SONIC, including but not limited to the data sampling strategy, network configuration, and rewards.

\paragraph{Results}
As shown in \reffig{fig:scaling-success-mpjpe}, \method{} achieves higher success rates and lower tracking errors than the retrained SONIC PD-target policy across all tested training-data scales, supporting the effectiveness and data efficiency of our design.
As training data increase, success approaches 100\% and the curves flatten near this ceiling, while tracking accuracy also improves, particularly on unseen motion categories.
From 4h onward, \method{} also tracks more accurately than the official SONIC checkpoint on both test sets.
As shown in \reffig{fig:scaling-qualitative}, \method achieves more accurate motion tracking than the official SONIC policy.
Together, these results support shifting the learning problem toward balanced execution while preserving motion fidelity through explicit tracking.
We hypothesize that this separation reduces the need for conservative execution that PD-target control may favor for stability, allowing the policy to learn increasingly transferable balancing capabilities as data scale.

\subsection{Tracking under External Force}

\label{sec:exp_interaction}

\begin{figure*}[t]
    \centering
    \includegraphics[width=\textwidth]{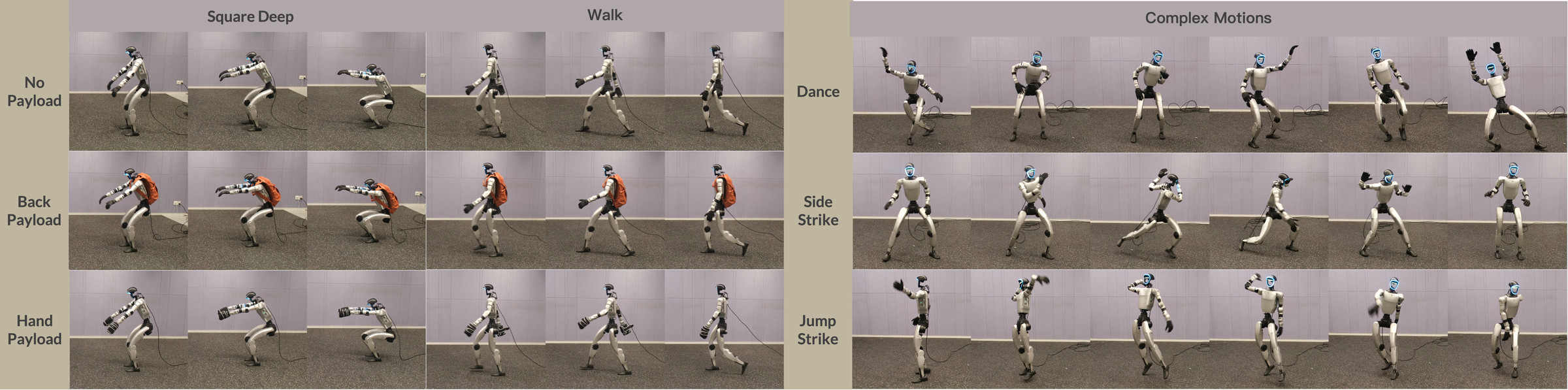}
    \caption{Qualitative hardware examples of \method on the G1 robot.
Left: deep squats and walking under different payloads; right: complex motions.
All motions are executed on the real robot using the same policy checkpoint.
Please refer to the supplementary video for more hardware results.}
    \label{fig:payload-qualitative}
\end{figure*}

\paragraph{Baseline}
We compare \method with FALCON~\cite{falcon} and ResMimic~\cite{resmimic} to examine tracking during external-force interaction.
FALCON adapts standing and locomotion to external forces, while ResMimic learns a residual around a general motion tracker.
We port the training code for FALCON and ResMimic from their official repositories and train and evaluate both methods on the same training set.
ResMimic trains a residual policy on top of a GMT policy.
We evaluate two variants, using either the officially released GMT checkpoint or a GMT policy trained from scratch on this experiment's training set.

\paragraph{Dataset}
To accommodate FALCON's control interface, we select standing and locomotion sequences from G1-retargeted LAFAN and AMASS Test.
The resulting set contains \textbf{217 motions: 209 locomotion and 8 stance sequences}.

\paragraph{Evaluation Protocol}
 We apply time-varying three-dimensional forces to both hands. During standing, the forces perturb the robot horizontally and vertically; during walking, the horizontal components oppose the commanded motion.
Force amplitudes are determined from the current arm Jacobians and joint torque limits, with $\alpha_g\in\{0,0.5,1\}$ scaling the generated forces to nominal zero, half, and full strength. Small random perturbations remain at $\alpha_g=0$.
We report upper-body MPJPE, upper-body MPJVE, root-velocity error, and success rate at each force level separately.
As shown in \reftab{tab:interaction-results}, \method achieves the lowest errors on all three tracking metrics at every force level, with comparable success rates.

\begin{table}[t]
    \centering
    \footnotesize
    \setlength{\tabcolsep}{2pt}
    \renewcommand{\arraystretch}{1.12}
    \begin{adjustbox}{max width=\linewidth,scale=0.88}
    \begin{tabular}{@{}llrrrr@{}}
        \toprule
        $\alpha_g$ & Method & \shortstack{Upper\\MPJPE $\downarrow$\\(mm)} & \shortstack{Upper\\MPJVE $\downarrow$\\(m/s)} & \shortstack{Root vel.\\error $\downarrow$\\(m/s)} & \shortstack{Success $\uparrow$\\(\%)} \\
        \midrule
        \multirow{4}{*}{$0.00$} & FALCON & 73.65 & 0.19 & 0.21 & 100.00 \\
         & ResMimic & 53.60 & 0.27 & 0.21 & 100.00 \\
         & ResMimic$^{*}$ & 39.29 & 0.27 & 0.19 & 100.00 \\
         & \method & 21.99 & 0.10 & 0.12 & 100.00 \\
        \midrule
        \multirow{4}{*}{$0.50$} & FALCON & 75.28 & 0.19 & 0.23 & 100.00 \\
         & ResMimic & 52.77 & 0.27 & 0.21 & 100.00 \\
         & ResMimic$^{*}$ & 42.17 & 0.27 & 0.21 & 100.00 \\
         & \method & 22.05 & 0.11 & 0.13 & 100.00 \\
        \midrule
        \multirow{4}{*}{$1.00$} & FALCON & 83.51 & 0.21 & 0.28 & 99.08 \\
         & ResMimic & 54.96 & 0.29 & 0.24 & 100.00 \\
         & ResMimic$^{*}$ & 54.01 & 0.30 & 0.27 & 99.08 \\
         & \method & 24.68 & 0.12 & 0.16 & 99.54 \\
        \bottomrule
    \end{tabular}
    \end{adjustbox}
    \caption{Comparison of \method with baselines. ResMimic$^{*}$ uses a GMT policy trained from scratch.}
    \label{tab:interaction-results}
\end{table}

\subsection{Ablation Studies}

\label{sec:exp_ablation}

We conduct three groups of ablations to examine the contribution of \method{}'s explicit decomposition and the effectiveness of its two components:

\begin{enumerate}
    \item \textbf{Explicit decoupling.} Compare explicitly decoupled and coupled formulations to assess the benefit of separating tracking from balancing.
    \item \textbf{Tracking Module.} Evaluate the tracking module without the learned residual to assess its ability to reproduce reference motions.
    \item \textbf{Balancing Execution.} Examine the effectiveness of learned balancing execution through its dependence on the tracking components and dynamics information.
\end{enumerate}

LAFAN contains a small number of long motion sequences.
To prevent a localized tracking failure from causing an entire long sequence to be counted as unsuccessful, we divide these sequences into clips of up to 10 seconds for evaluation.

\subsubsection{Decoupled versus Coupled Control}

\label{sec:exp_force_spectrum}

\paragraph{Baseline}
We use BeyondMimic~\cite{beyondmimic} as the coupled-control baseline and \method as the decoupled formulation, comparing position-target tracking with explicit analytic tracking plus a learned balancing residual.
To accommodate the larger training set, both methods use the same adaptive motion sampling strategy and an enlarged policy network.
Apart from the policy prediction targets, we keep their training settings identical, including rewards, domain randomization, network architecture, training budget, and maximum episode length, to ensure a fair comparison.
Both methods are trained for 10,000 updates on LAFAN. 
During training, we introduce both attached payloads and environmental disturbances so that the controllers learn to track motions under external perturbations.

\paragraph{Evaluation Protocol}
We evaluate on LAFAN and assess transfer to BONES-SEED test-content and test-repetition~\cite{sonic} using the same checkpoints without fine-tuning.
At test time, payloads and environmental disturbances are evaluated separately: G denotes ground tracking without additional loads, P covers a 5 kg backpack and 2 kg in each hand, and I applies horizontal root-velocity perturbations.

\reftab{tab:force-spectrum} reports success rate together with MPJPE and MPJVE.
Both errors are computed on the exact set of motions successfully completed by both methods within each condition, with equal weight per motion.
For P, we first compute each payload condition separately and then average their results with equal weight.

\begin{table*}[tp]
    \centering
    \footnotesize
    \setlength{\tabcolsep}{3.5pt}
    \renewcommand{\arraystretch}{1.12}
    \begin{adjustbox}{max width=\linewidth,scale=0.95}
    \begin{tabular}{@{}ll*{9}{r}@{}}
        \toprule
        & & \multicolumn{3}{c}{G: No additional load} & \multicolumn{3}{c}{P: Backpack and hand payloads} & \multicolumn{3}{c}{I: Environmental disturbance} \\
        \cmidrule(lr){3-5}\cmidrule(lr){6-8}\cmidrule(lr){9-11}
        Dataset & Method & \shortstack{Success $\uparrow$\\(\%)} & \shortstack{MPJPE $\downarrow$\\(mm)} & \shortstack{MPJVE $\downarrow$\\(m/s)} & \shortstack{Success $\uparrow$\\(\%)} & \shortstack{MPJPE $\downarrow$\\(mm)} & \shortstack{MPJVE $\downarrow$\\(m/s)} & \shortstack{Success $\uparrow$\\(\%)} & \shortstack{MPJPE $\downarrow$\\(mm)} & \shortstack{MPJVE $\downarrow$\\(m/s)} \\
        \midrule
        \multirow{2}{*}{LAFAN} & BeyondMimic (coupled) & 97.67 & 34.06 & 0.29 & 95.75 & 38.96 & 0.32 & 96.51 & 37.66 & 0.38 \\
         & \method & \textbf{98.95} & \textbf{26.59} & \textbf{0.24} & \textbf{97.73} & \textbf{29.68} & \textbf{0.27} & \textbf{97.67} & \textbf{30.22} & \textbf{0.33} \\
        \midrule
        \multirow{2}{*}{\shortstack[l]{BONES-SEED\\test-content}} & BeyondMimic (coupled) & 92.15 & 38.97 & 0.27 & 87.06 & 42.71 & 0.29 & 90.30 & 42.07 & 0.36 \\
         & \method & \textbf{93.07} & \textbf{30.66} & \textbf{0.23} & \textbf{89.73} & \textbf{34.71} & \textbf{0.26} & \textbf{92.18} & \textbf{33.65} & \textbf{0.32} \\
        \midrule
        \multirow{2}{*}{\shortstack[l]{BONES-SEED\\test-repetition}} & BeyondMimic (coupled) & 92.25 & 36.68 & 0.25 & 87.16 & 41.03 & 0.28 & 90.69 & 40.28 & 0.35 \\
         & \method & \textbf{93.64} & \textbf{30.24} & \textbf{0.23} & \textbf{89.15} & \textbf{34.10} & \textbf{0.26} & \textbf{92.48} & \textbf{32.78} & \textbf{0.32} \\
        \bottomrule
    \end{tabular}
    \end{adjustbox}
    \caption{Experimental results for \method and BeyondMimic (coupled).
LAFAN is the training set, while BONES-SEED test-content and test-repetition~\cite{sonic} are test sets.
Better results are shown in bold.
}
    \label{tab:force-spectrum}
\end{table*}

\begin{table*}[t]
    \centering
    \footnotesize
    \setlength{\tabcolsep}{3pt}
    \renewcommand{\arraystretch}{1.12}
    \begin{adjustbox}{max width=\textwidth,scale=0.95}
    \begin{tabular}{@{}lll*{6}{r}@{}}
        \toprule
        \multirow{2}{*}{Variant} & \multirow{2}{*}{\shortstack{Tracking Module\\components}} & \multirow{2}{*}{\shortstack{Observations\\(actor ID)}} & \multicolumn{3}{c}{LAFAN} & \multicolumn{3}{c}{BONES-SEED test-repetition} \\
        \cmidrule(lr){4-6}\cmidrule(lr){7-9}
         & & & \shortstack{Success $\uparrow$\\(\%)} & \shortstack{MPJPE $\downarrow$\\(mm)} & \shortstack{MPJVE $\downarrow$\\(m/s)} & \shortstack{Success $\uparrow$\\(\%)} & \shortstack{MPJPE $\downarrow$\\(mm)} & \shortstack{MPJVE $\downarrow$\\(m/s)} \\
        \midrule
        \method{} (full) & ID + PD & full & 95.93 & 32.91 & 0.30 & \textbf{90.86} & \textbf{43.16} & \textbf{0.25} \\
        wo PD & ID & full & 89.41 & 44.52 & 0.36 & 88.64 & 50.82 & 0.27 \\
        wo ID & PD & full & 91.97 & 38.04 & 0.34 & 88.64 & 48.50 & 0.28 \\
        wo ID (FF + obs) & PD & wo all ID & 91.15 & 39.94 & 0.36 & 86.65 & 53.62 & 0.32 \\
        Policy only (torque) & None & wo all ID & 88.82 & 42.12 & 0.38 & 83.94 & 55.12 & 0.31 \\
        Policy only (PD target) & None & wo all ID & 90.69 & 41.23 & 0.36 & 84.90 & 57.15 & 0.33 \\
        \method{} (PD target) & ID + PD & full & 94.88 & 35.91 & 0.30 & 90.43 & 46.06 & 0.27 \\
        \midrule
        wo actor ID observations & ID + PD & wo all ID & \textbf{96.16} & \textbf{32.19} & 0.30 & 89.17 & 48.61 & 0.27 \\
        wo ID acceleration & ID + PD & wo acceleration & \textbf{96.62} & \textbf{32.23} & 0.30 & 88.77 & 48.40 & 0.28 \\
        wo ID Coriolis & ID + PD & wo Coriolis & \textbf{96.04} & \textbf{31.77} & 0.30 & 90.46 & 43.75 & 0.26 \\
        wo ID gravity & ID + PD & wo gravity & \textbf{96.04} & \textbf{32.45} & 0.30 & 90.00 & 46.16 & 0.26 \\
        \bottomrule
    \end{tabular}
    \end{adjustbox}
    \caption{Component ablations on LAFAN and BONES-SEED test-repetition~\cite{sonic}.
Bold values outperform full \method on LAFAN or are best across variants on BONES-SEED test-repetition.}
    \label{tab:balancing-ablation}
\end{table*}

\method achieves higher success and lower position and velocity errors on LAFAN and both BONES-SEED test splits under all G/P/I conditions.
Although the two methods achieve similar success rates, coupled control incurs substantially larger tracking errors.
One possible explanation is that the coupled policy favors more conservative control outputs, maintaining stability at the expense of tracking accuracy.
These results show that \method{}'s gains in tracking accuracy and motion completion extend to unseen motion datasets under payloads and external disturbances.

\begin{figure}[t]
    \centering
    \includegraphics[width=0.95\linewidth]{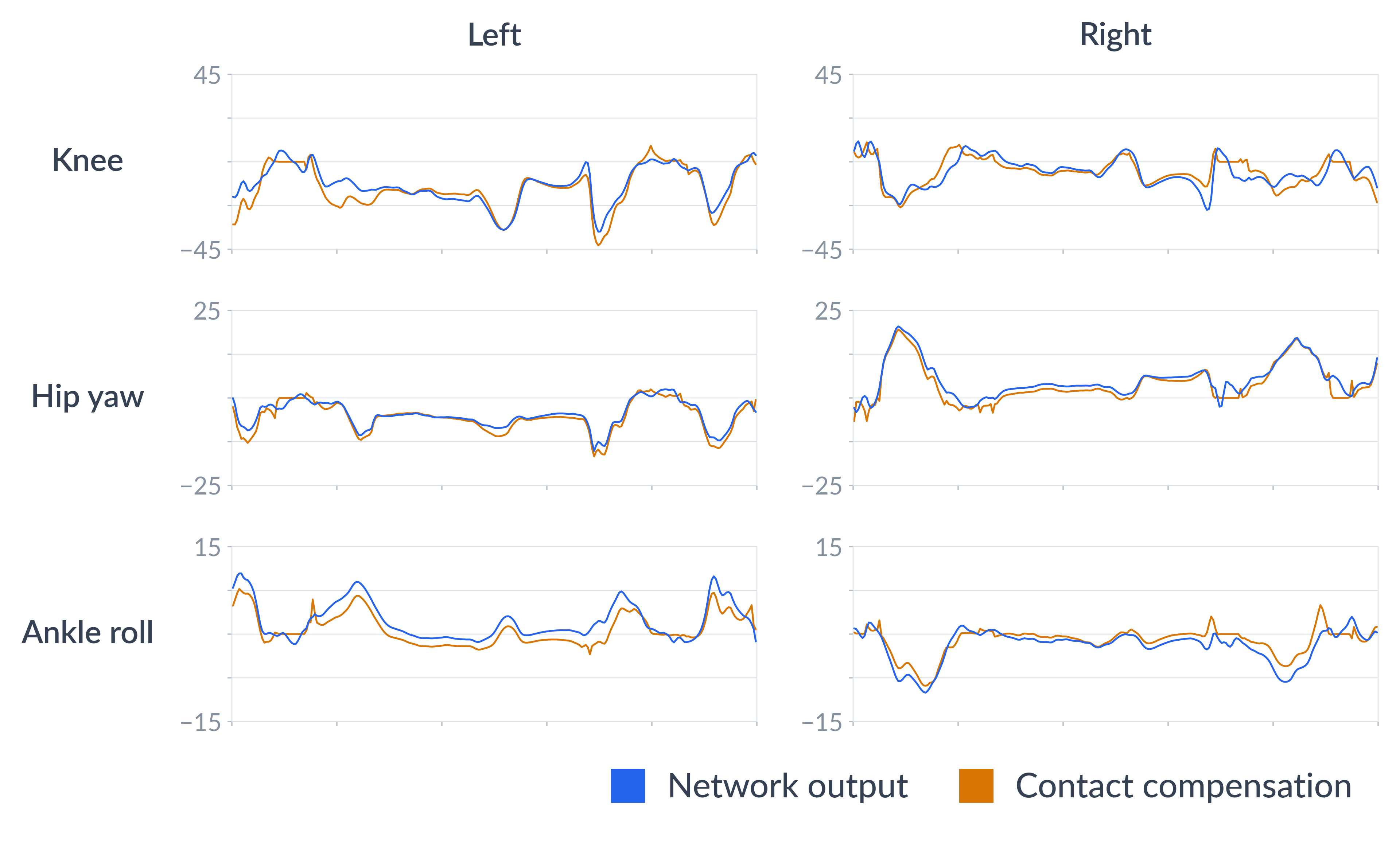}
    \caption{Learned residual and contact-compensation torques during a LAFAN motion segment executed by \method with ground contact in simulation.
All panels show the same time interval from the same execution.
The horizontal axis is time (s), and the vertical axis is joint torque ($\mathrm{N\!\cdot\!m}$).
Blue denotes the network's residual torque $\boldsymbol\tau_{\mathrm{bal}}$; orange denotes the contact-compensation torque $-\mathbf S\sum_k\mathbf J_k^\top\boldsymbol\lambda_k$, obtained by substituting simulator contact wrenches and their Jacobians into the environmental wrench term in Eq.~\eqref{eq:method-actuated-dynamics}.}
    \label{fig:residual-contact-compensation}
\end{figure}

\paragraph{Interpretation of the learned residual}
\reffig{fig:residual-contact-compensation} compares the learned residual with contact-compensation torques for selected joints during a single LAFAN motion segment with ground contact in simulation.
We compute the contact-compensation torque by substituting simulator contact wrenches and their corresponding Jacobians into the environmental wrench term $-\mathbf S\sum_k\mathbf J_k^\top\boldsymbol\lambda_k$ in Eq.~\eqref{eq:method-actuated-dynamics}.
The agreement in amplitude and temporal variation is consistent with the residual's intended role in contact compensation during this segment.

\Needspace{5\baselineskip}
\subsubsection{Effectiveness of the Tracking Module}

\label{sec:exp_fixed_root}

\paragraph{Baseline}
We compare the tracking module with PD controllers at different gains, keeping the damping ratio fixed as stiffness increases.
We do not increase stiffness further because a stiffness factor of 3 already frequently drives the commanded torques to the joint torque limits.

\paragraph{Evaluation Protocol}
We ask whether the tracking module alone can reproduce the reference motion.
All comparisons use a fixed root in mid-air, with gravity enabled and no contact.
We disable the learned residual to isolate the tracking torque.
Each controller applies its torque to the same robot from the same initial state while tracking the LAFAN references.
We evaluate MPJPE, MPJVE, and MPJAE to measure position, velocity, and acceleration tracking.

\begin{table}[t]
    \centering
    \footnotesize
    \setlength{\tabcolsep}{3pt}
    \renewcommand{\arraystretch}{1.12}
    \begin{adjustbox}{max width=\linewidth,scale=0.85}
    \begin{tabular}{@{}llrrr@{}}
        \toprule
        Controller & \shortstack{$K_p/K_d$\\scale} & \shortstack{MPJPE $\downarrow$\\(mm)} & \shortstack{MPJVE $\downarrow$\\(m/s)} & \shortstack{MPJAE $\downarrow$\\(m/s$^2$)} \\
        \midrule
        \multirow{3}{*}{PD} & $1 / 1$ & 63.19 & 0.31 & 4.32 \\
         & $2 / \sqrt{2}$ & 34.24 & 0.25 & 4.02 \\
         & $3 / \sqrt{3}$ & 23.80 & 0.21 & 3.75 \\
        \midrule
        FF only & $0 / 0$ & 171.73 & 0.40 & 4.47 \\
        Tracking module & $1 / 1$ & 4.45 & 0.04 & 0.89 \\
        \bottomrule
    \end{tabular}
    \end{adjustbox}
    \caption{Comparison of tracking module with PD variants.}
    \label{tab:fixed-root}
\end{table}

\reftab{tab:fixed-root} summarizes the tracking errors.
The tracking module achieves 4.45 mm MPJPE. These results show that the tracking module achieves substantially higher tracking accuracy than PD feedback alone, even with amplified gains.
The much larger error with FF alone (171.73 mm MPJPE) indicates the limited dynamic feasibility of retargeted human motions, underscoring the need for position feedback in the tracking module.

\subsubsection{Effectiveness of Balancing Execution}

\label{sec:exp_balancing}

Through action and observation ablations, we examine the roles of the Tracking Module and dynamics information in balanced execution.

\paragraph{Baseline}
We compare full \method{} with ablated variants, as summarized in \reftab{tab:balancing-ablation}.
Training is conducted on LAFAN using matched networks, rewards, and domain randomization.
Training uses ground support with the same BeyondMimic domain randomization, including pushes, without added payloads or the mixed force scenarios used in Section~\ref{sec:exp_force_spectrum}.

\paragraph{Evaluation Protocol}
We evaluate complete motions on LAFAN and BONES-SEED test-repetition~\cite{sonic}, with MPJPE and MPJVE computed over all motions.

\paragraph{Action design}
Full \method{} achieves higher success and lower MPJPE and MPJVE than the action variants on both datasets.
Removing both ID feedforward and actor ID observations (wo ID (FF + obs)) further reduces success and increases tracking errors compared with removing ID feedforward alone (wo ID), indicating that dynamics observations remain useful when only reference-motion PD feedback is retained.
With the full Tracking Module, torque prediction improves tracking accuracy over PD-target prediction on BONES-SEED test-repetition, although the gain is smaller than the gain over the torque-only policy.
Without the Tracking Module, however, Policy only (PD target) achieves higher success than Policy only (torque) on both datasets, while its tracking errors are higher on BONES-SEED test-repetition.
These results support pairing torque prediction with analytic tracking to separate motion following from balanced execution; they do not establish a general advantage of torque actions without a tracker.
The comparisons with matched training budgets also support retaining PD feedback.

\paragraph{Observation design}
Removing all or any one of the actor ID components improves success and lowers MPJPE on LAFAN, which shares the training motion set.
MPJVE does not improve consistently: only removing the Coriolis component lowers it.
This is because removing these observations makes the policy more prone to overfitting.
On the held-out BONES-SEED test-repetition set, every observation ablation reduces success and increases both MPJPE and MPJVE.
Thus, improved fitting on LAFAN does not translate into better transfer.
This pattern supports our interpretation that explicit tracking components provide useful context for learning transferable balancing execution, encouraging the policy to account for the analytic tracker's contribution rather than relearn motion-specific tracking compensation.

\section{Conclusion}
\label{sec:conclusion}

We presented \method, a humanoid whole-body control framework that separates model-based motion tracking from learned support control.
By explicitly supplying nominal tracking torques through floating-base inverse dynamics and joint feedback, \method reframes policy learning as achieving balanced execution across diverse reference motions.
Across the tested training-data scales, \method achieves higher success rates and lower tracking errors than the adapted SONIC baseline on both BONES-SEED test sets.
Further experiments demonstrate improved cross-dataset transfer and tracking under payloads and external disturbances, while ablations highlight the contribution of inverse-dynamics observations to generalization beyond the training motions.
Tests on the Unitree G1 demonstrate diverse motion execution and tracking under attached payloads.
Together, these results support explicitly modeling nominal motion dynamics as a foundation for learning transferable support capabilities while preserving tracking accuracy.
Future work will explore extending this decomposition to a broader range of whole-body physical interaction tasks.

{\small
\begingroup
\let\url\nolinkurl
\let\footnotesize\small
\let\originalthebibliography\thebibliography
\renewcommand{\thebibliography}[1]{\originalthebibliography{#1}\interlinepenalty=10000\relax}
\bibliographystyle{IEEEtran}
\bibliography{main}
\endgroup
}

\end{document}